# ECG Heartbeat classification using deep transfer learning with Convolutional Neural Network and STFT technique

Minh Cao[1], Tianqi Zhao[2], Yanxun Li[2], Wenhao Zhang[2], Peyman Benharash[3], Ramin Ramezani[2]

[1] Bioengineering Dept, University of California, Los Angeles, Los Angeles CA 90095, USA
[2] Computer Science Dept, University of California, Los Angeles, Los Angeles CA 90095, USA
[3] David Geffen School of Medicine Dept, University of California, Los Angeles, Los Angeles CA 90095, USA

**Abstract.** Electrocardiogram (ECG) is a simple non-invasive measure to identify heart-related issues such as irregular heartbeats known as arrhythmias. While artificial intelligence and machine learning is being utilized in a wide range of healthcare related applications and datasets, many arrhythmia classifiers using deep learning methods have been proposed in recent years. However, sizes of the available datasets from which to build and assess machine learning models is often very small and the lack of well-annotated public ECG datasets is evident. In this paper, we propose a deep transfer learning framework that is aimed to perform classification on a small size training dataset. The proposed method is to fine-tune a general-purpose image classifier ResNet-18 with MIT-BIH arrhythmia dataset in accordance with the AAMI EC57 standard. This paper further investigates many existing deep learning models that have failed to avoid data leakage against AAMI recommendations. We compare how different data split methods impact the model performance. This comparison study implies that future work in arrhythmia classification should follow the AAMI EC57 standard when using any including MIT-BIH arrhythmia dataset.

**Keywords:** Deep transfer learning, ECG, Arrhythmia classification

## 1 Introduction

Electrocardiogram (ECG) is a simple non-invasive measure to identify heart-related issues such as irregular heartbeats known as arrhythmias. Even though sometimes being observed in healthy people, arrhythmias can develop life-threatening cardiac diseases. Manual inspection on ECG signals to identify arrhythmia can be time consuming and error-prone [1,2]. Due to the capability of learning complex representation, there have been major developments in utilizing deep learning methods for automatic ECG-based arrhythmia diagnosis [1-13].

Deep learning methods generally require a large amount of training data. While a well-annotated ECG data for arrhythmia detection is limited [8], resorting to transfer learning techniques in which a pre-trained image classifier is used can be warranted. There have been recent attempts in using transfer learning framework with MIT-BIH dataset to develop arrhythmia diagnosis models [1]. MIT-BIH arrhythmia database is the most widely used dataset in developing and evaluating ECG-based arrhythmia models [3,6]. The Association for the Advancement of Medical Instrumentation (AAMI) has developed a standard for testing and reporting performance results of diagnostic models for arrhythmia classification (ANSI/AAMI EC57:1998/(R)2008). Unfortunately, we identified that several deep learning models failed to follow the standard and produced misleading and unreliable results with the MIT-BIH Arrhythmia database, reporting near perfect classification performance (e.g., >90% overall accuracy) [1,3,13]. When evaluated the models according to AAMI recommendations, the classification performance drops significantly. Hence, the results of such models should be considered and interpreted for clinical decision making with a grain of salt.

To address the issues and challenges mentioned above, we propose an end-to-end ECG classification framework by leveraging the transfer learning framework. We use the Short-Term Fourier Transform (STFT) technique to convert 1D ECG signals to 2D time-frequency domain data which can later be used in the 2D CNN pretrained models such as VGGs and ResNets.

The main contributions of this paper are summarized as follows:

- Propose an end-to-end ECG classification framework that can leverage the learning power of existing pre-trained 2D CNN models.
- Demonstrate how the choice of samples in MIT-BIH dataset significantly impacts the deep learning model performance.
- Highlight the unreliable and biased model evaluation in current literature on ECG classification with deep learning methods.

## 2  Related Studies

In the literature, the ECG analysis generally consists of the following steps: 1) ECG signal preprocessing and noise attenuation, 2) heartbeat segmentation, 3) feature extraction, and 4) learning/classification [2].

The first three steps have been widely studied and discussed in the literature [9-14]. In this section, we only review a selection of these methods due to page limit. For example, Omid Sayadi et al. proposed a modified extended Kalman filter structure which can be used not only for denoising the ECG signal, but also for compression [9]. C Li et al. presented a heartbeat segmentation algorithm based on wavelet transforms (WT's), which can detect QRS complex (see Figure 1) from high P or T waves even with the existence of serious noise or drift [10]. As for feature extraction, Chun-Cheng

Lin et al. proposed an automatic heartbeat classification system for arrhythmia classification based on normalized RR intervals (i.e., interval between two successive R waves) and morphological features derived from wavelet transform and linear prediction modeling [14].

Machine learning models are widely used for arrhythmia classification in the literature [2,3,5,7,8,13,14,15]. Mi Hye Song et al. proposed a support vector machine-based classifier with reduced features derived by linear discriminant analysis [5]. Inspired by the success of Hidden Markov Model (HMM) in modeling speech waveforms for automatic speech recognition, D A Coast et al. applied HMM method in ECG arrhythmia analysis. The model can combine the temporal information and statistical knowledge of the ECG signal in one single parametric model [15]. Awni Y. Hannun et al. proposed an end-to-end deep learning approach which directly takes raw ECG signal as input and produces classifications without feature engineering or feature selection [8]. Mousavi, Sajad et al. proposed an automatic ECG-based heartbeat classification approach by utilizing a sequence-to-sequence deep learning method to automatically extract temporal and statistical features of the ECG signals [16].

Our work differs from the studies in 2-fold: 1) it leverages the Short-term Fourier Transform (STFT) to convert 1D ECG signal into 2D time-frequency domain data. Therefore, it is feasible to apply pre-trained 2D Convolution Neural Network in arrhythmia analysis; 2) it is evaluated using MIT-BIH dataset with "*inter-patient*" training/testing split paradigm detailed in [17].

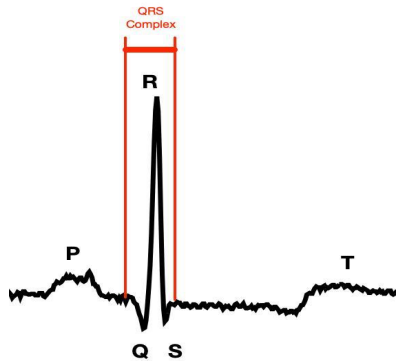

**Fig. 1.** A typical heartbeat ECG signal contains P, Q, R, S, and T waves. A QRS complex is a combination of Q,R,S waves. See Table A.1 for a complete list of the rhythm.

## 3 Datasets

Similar to the majority of the arrhythmia analysis studies, this study develops the model on MIT-BIH Arrhythmia dataset [6]. The dataset includes 48 half-hour excerpts of two-channel ambulatory ECG recordings collected from 47 patients at 360 Hz. The dataset was annotated at heartbeat level by two or more cardiologists independently. 14 original heartbeat types are consolidated into 5 groups according to AAMI recommendation, shown in Figure 2.

Table 1. Heartbeat distribution by classes of the raw data, intra-patient split, and inter-patient split.

| Set | Heartbeat types (AAMI EC57 standard) | | | | | |
|---|---|---|---|---|---|---|
| | N | S | V | F | Q | Total |
| Full MIT-BIH set | 90,631 | 2,781 | 7,236 | 803 | 8,043 | 109,494 |
| Intra-patient split | | | | | | |
| Training (80 % split) | 72,471 | 2,223 | 5,789 | 642 | 6,431 | 87,756 |
| Testing (20 % split) | 18,118 | 556 | 1,447 | 161 | 1,608 | 21,890 |
| Inter-patient split | | | | | | |
| Training (DS1 in [17]) | 45,866 | 944 | 3,788 | 415 | 8 | 51,021 |
| Testing (DS2 in [17]) | 44,259 | 1,837 | 3,221 | 388 | 7 | 49,712 |

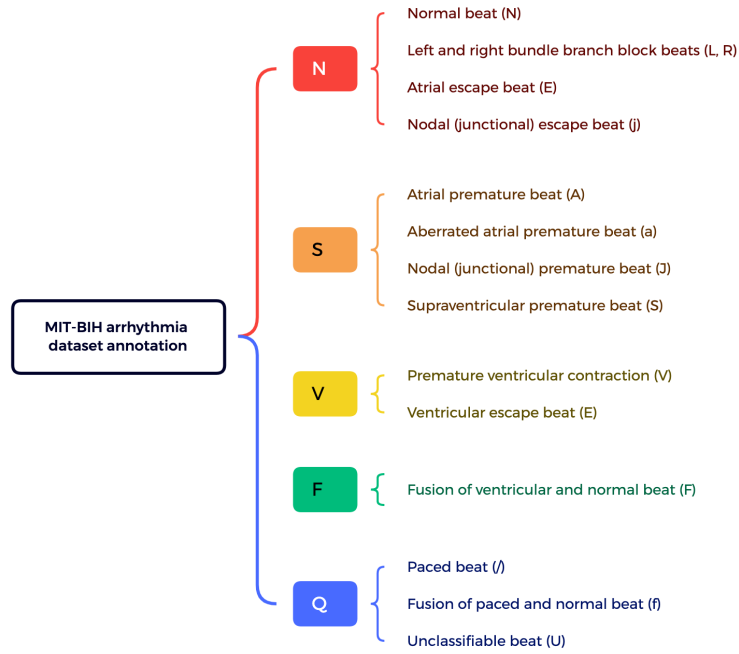

Fig. 2. Heartbeat annotations in MIT-BIH dataset according to AAMI EC 57. The consolidated classes are N, S, V, F, Q.

There are two ways to split the dataset into training and testing sets, inter-patient paradigm versus intra-patient paradigm. The intra-patient paradigm creates the training/testing dataset by randomly choosing heartbeat samples. In this paradigm, the heartbeat samples from the same patient might exist in both training and testing dataset. Therefore, the testing data might influence the model training. We argue that this data split paradigm can result in unreliable results. Models developed under intra-patient split paradigm should be reconsidered and re-evaluated for clinical decision making; overall, intra-patient approach should be highly discouraged [4,17]. On the other hand, in inter-patient paradigm the training/testing datasets are created from different patients [17,18]. Hence, inter-patient split avoids the information leakage issue existed in its counterpart method. In the inter-patient split, the MIT-BIH dataset is divided into two datasets (DS1 and DS2), identified by the patient IDs: DS1 = {101, 106, 108, 109, 112, 114, 115, 116, 118, 119, 122, 124, 201, 203, 205, 207, 208, 209, 215, 220, 223, 230} and DS2 = {100, 103, 105, 111, 113, 117, 121, 123, 200, 202, 210, 212, 213, 214, 219, 221, 222, 228, 231, 232, 233, 234} proposed by Chazal et al. [17]. DS1 is used for model training (training set) and DS2 is used for model evaluation (test set). Patient 102, 104, 107, 217 are excluded in inter-patient split.

## 4     Methods

In this section, we describe the proposed work in two steps: 1) building an end-to-end transfer learning framework with STFT and ResNet18, and 2) investigating how the inter-patient split and intra-patient split of MIT-BIH dataset impact the performance of a series of models presented in the literature [1,3,13].

### 4.1     Preprocessing

The raw MIT-BIH data firstly goes through a high pass filter (> 0.5 Hz) to remove baseline constant signal. Moving average is applied to remove base drift. Chebyshev type I $4^{th}$ order filter and bandwidth 6-18 Hz coupled with Shannon energy filters are used to find the R peak. Then the ECG recordings are segmented into a set of heartbeats with the length of 1.2 RR interval (the time between two consecutive peaks).

In our proposed approach, we use pretrained 2D CNN models (ResNet18) which requires the input data to be in the format of 2D images. Therefore, Short-Term Fourier Transform (STFT) is used to obtain 2D time-frequency spectrograms of the digitized 1D ECG recordings for capturing the frequency variations [19,20]. The 2D time-frequency spectrograms for each point in the signal is computed by [20],

$$STFT\{x[n]\} = X(x, \omega) = \sum_{n=-\infty}^{\infty} x[n] w[n-m] e^{-j\omega n} \quad (1)$$

Where $x[n]$ is the signal which is sampled at 360 Hz and $w[n-m]$ is the moving window (e.g., Hanning window or Guassian window). We suggest using Hanning window with size 512. The resulting 2D spectrograms are in dimension of 224 X 224. The STFT transformation is performed using Python library librosa (version 0.9.1).

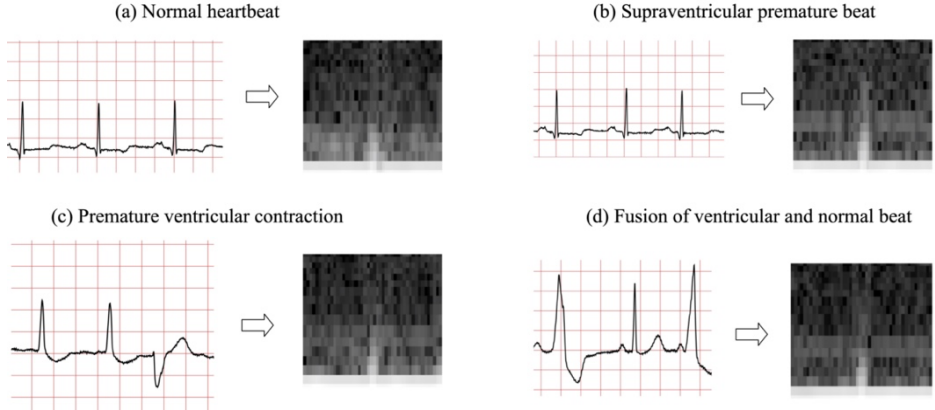

**Fig. 3.** ECG grey-scaled spectrograms of the 4 class in MIT BIH dataset.

The class distribution in MIT-BIH dataset is highly imbalanced (the majority class N takes 89% of the entire dataset, see Table 1). In this study, we only consider the heartbeat type N, S, V, F, and exclude class Q due to limited samples (n=15). Over-sampling and under-sampling techniques are explored in this study to construct equalized class representation. Note that data sampling methods are only applied on training dataset DS1. We applied oversampling after STFT is performed. Image rotation, flip, and adding Gaussian noise are used to create the artificial data samples.

### 4.2 Arrhythmia Classifier using Transfer Learning

Inspired by the application of using pretrained CNN classifiers (e.g., ResNet18, ResNet50, etc.) to build predictive models in lung CT scans, we explored the feasibility of using such classifiers in Arrhythmia classification [21]. ResNet18 (Figure A1) is used to classify ECG recordings into 4 classes listed in Table A1. The dimension of the input data is adjusted to 224 × 224 × 1 for ResNet 18. A fully connected layer at the end of ResNet18 is adjusted to predict 4 classes. To classify Arrhythmia, the pretrained ResNet18 network is fine-tuned using the preprocessed DS1 dataset from inter-patient split paradigm. Then the retrained ResNet18 is evaluated using DS2.

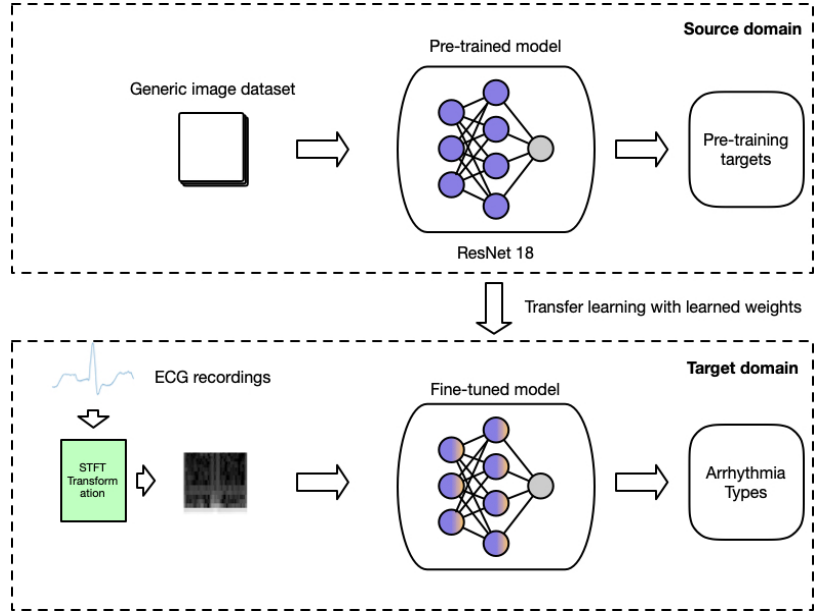

**Fig. 4.** Visualization of transfer learning in this work. The pretrained models are developed on generic image dataset. There are wide choices of existing pretrained models such as ResNet18 and VGGs. The pretrained models are then fine-tuned with task-specific data, i.e., 2D ECG data in time-frequency domains transformed from 1D ECG waveform recordings. We suggest using pretrained ResNet 18 for classification.

The training parameters opted for the transfer learning-based model are: i) using Adam optimizer, ii) batch size is 500, iii) model is trained up to 20 epochs, iv) learning rate is .0001. The metrics used for evaluation are precision, recall, and accuracy.

### 4.3 Investigation of how the choice of Intra-patient split versus Inter-patient split paradigm impact model performance

In this section, we investigate how the inter-patient split and intra-patient split methods impact the performance of several state-of -the-art deep learning models in the literature [1,3]. Eduardo Luz et al. studied the impact of these two data split paradigms [4] on machine learning models such as SVM and shallow neural networks. However, that study was published in 2011 and did not include deep learning models. In our study, we implemented two state-of-the-art deep learning models: Convolutional Neural Networks and its variant described in [1,3]. Kachuee et al. only evaluates their arrhythmia classifier on intra-patient split [1]. However, we re-implemented[1] these deep

---

[1] Our implementation can be found at *https://github.com/wenhaoz-fengcai/ECG-Arrhythmia-Detection-DL*.

learning models and tested the models using both intra-patient and inter-patient split paradigms.

## 5 Results

The performance of the proposed transfer learning framework model is presented in Table 2. We also report the comparison study of existing models in the literature and show the performance difference between inter-patient paradigm and intra-patient paradigm in Table 2 and Table 3, respectively.

**Table 2.** Performance comparison of deep learning models with *inter-patient split paradigm*. The metrics reported are overall accuracy, precision (Pre), and recall (Rec). Note that the first model was not tested using inter-patient split paradigm in the original paper. The results obtained here are from our re-implementations. The best scores are bold-faced in each column.

| Work | Accuracy (%) | Arrhythmia types | | | | |
|---|---|---|---|---|---|---|
| | | N (n=44,259) | S (n=1,837) | V (n=3,221) | F (n=388) | Q (n=7) |
| | | Pre/Rec | Pre/Rec | Pre/Rec | Pre/Rec | Pre/Rec |
| Kachuee[1] | 81.2 | 94.4/84.5 | 0.0/0.0 | 30.9/**92.4** | 1.0/**1.3** | 0.0/0.0 |
| Romdhane[3] | 62.1 | **95.6**/64.0 | 0.0/0.0 | 12.7/79.3 | 0.0/0.0 | 0.0/0.0 |
| **Proposed method** | **90.8** | 95.3/**95.1** | **13.0**/**9.0** | **68.2**/88.4 | **1.3**/0.3 | N/A |

It is worth noting that authors in [1] tried to mitigate the imbalance problem even in the "test set" which is contrary to machine learning practice; mitigation occurs during training. Also, in [1] the number of test samples in F class is equal to the entire class which may explain the high precision and recall.

**Table 3.** Performance of deep learning model re-implementations with *intra-patient split paradigm*. The reported metric are the overall accuracy, precision (Pre), and recall (Rec) of our implementation. The numbers in paratheses are results reported in the literature.

| Work | Accuracy (%) | Arrhythmia types | | | | |
|---|---|---|---|---|---|---|
| | | N (n=18,118) | S (n=556) | V (n=1,447) | F (n=161) | Q (n=1,608) |
| | | Pre/Rec | Pre/Rec | Pre/Rec | Pre/Rec | Pre/Rec |
| Kachuee[1] (reported in literature) | 93.1 (93.4) | 98.4/94.3 (84.3/97.0) | 38.1/82.6 (98.9/89.0) | 96.6/82.3 (95.0/96.0) | 26.6/93.8 (100.0/86.0) | 98.3/92.6 (100.0/98.0) |
| Romdhane[3][2] | 82.7 | 82.8/99.9 | 0.0/0.0 | 42.9/0.4 | 0.0/0.0 | 0.0/0.0 |

---

[2] [3] authors claimed that their model was evaluated using inter-patient split paradigm even if it is based on intra-patient split.

Table 2 shows the proposed ResNet18 model with data augmentation achieves best overall accuracy, best recall in normal (N) class, best precisions in arrhythmia (S, V, F) classes. It is worth mentioning that the first model [1] in Table 2 was only tested using intra-patient split paradigm in the original paper. In addition, the results from our re-implementation in study [3] are not similar to the reported numbers. We notice that the evaluation procedure is based on intra-patient split even though the authors claimed that their model was tested using inter-patient split method [3]. Table 3 presents the model testing results from our implementation as well as the reported performance in the literature. There is a significant performance drop once deep learning models are trained using inter-patient split instead of intra-patient split for deep learning models described in [1,3]. Huang et al. only evaluated their model on heartbeat type N, S, V, hence the testing results under F and Q are not available.

## 6   Discussion and Conclusion

We proposed an end-to-end ECG classification framework using 2D CNN classifiers. By transforming the 1D ECG waveforms into 2D frequency-time spectrogram using Short-Term Fourier Transform, the proposed framework provides the opportunity of integrating the general purpose pre-trained 2D CNN models (e.g., VGG-16, Efficient Net, etc) for arrhythmia detection. The proposed method achieves better overall accuracy compared with deep learning models described in [1,3].

Our second contribution is to demonstrate how the choice of samples in MIT-BIH dataset significantly impacts the deep learning model performance. We re-implemented two deep learning models for arrhythmia detection in [1,3], and then tested these models following the AAMI recommendation using the inter-patient data split. We observe that the model evaluation using intra-patient split generates better results compared with the testing results using inter-patient paradigm. However, we argue that the testing set of intra-patient paradigm is susceptible to contamination and is highly likely to have included samples from the same patients appeared in the training set. Therefore, the intra-patient split paradigm is more likely to generate inflated and biased results compared with inter-patient split paradigm.

In addition, there is a lack of consistency regarding the usage of MIT-BIH dataset for arrhythmia classification. For example, study [13] only includes data samples from 14 patients out of 47 in the MIT-BIH dataset. Meanwhile, classifiers in studies [5,13,18,22] only predict a subset of heartbeat types. In [13], only normal beat (NOR), left bundle branch block beat (LBB), right bundle branch block beat (RBB), pre-mature ventricular contraction beat (PVC), atrial premature contraction beat (APC) are included in the analysis. Moreover, there is a lack of standard reporting in the arrhythmia classification literature. For example, [1,3] evaluate their models using precision, recall, and overall accuracy, whereas [4,5] reported their models performance using specificity and sensitivity.

With the intention of building robust and unbiased arrhythmia classifiers, we highly suggest that practitioners follow the correct practice of splitting the training and testing data to avoid any possible information leakage. Moreover, we are calling for more transparency of data preprocessing and model development, along with a standard of model evaluation. In such a manner, the research community can reproduce and verify the results.

# Appendix

**Table A1.** A complete list of the rhythm considered in this study. The Q class is ignored in this study due to the limited number of samples (n=15). The table also shows the class distribution with inter-patient split paradigm. **N**, **S**, **V**, **F** represent the **N**ormal heartbeat, **S**upraventricular premature beat, premature **V**entricular contraction, **F**usion of ventricular and normal beat, respectively.

| Class | Heartbeats in DS1 | Heartbeats in DS2 | ECG Example |
|---|---|---|---|
| N | 45,866 | 44,259 | 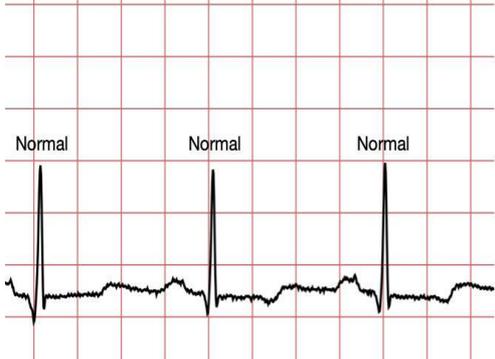 |
| S | 944 | 1,837 | 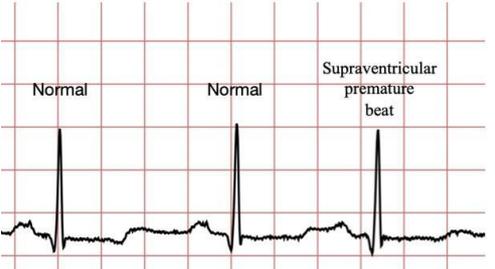 |
| V | 3,788 | 3,221 | 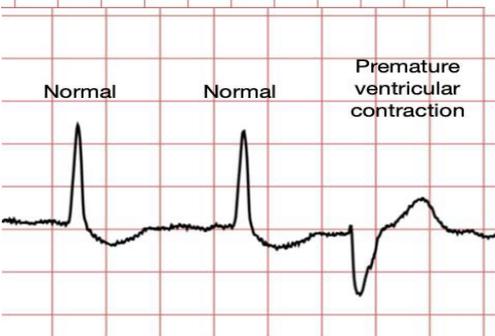 |
| F | 415 | 388 | 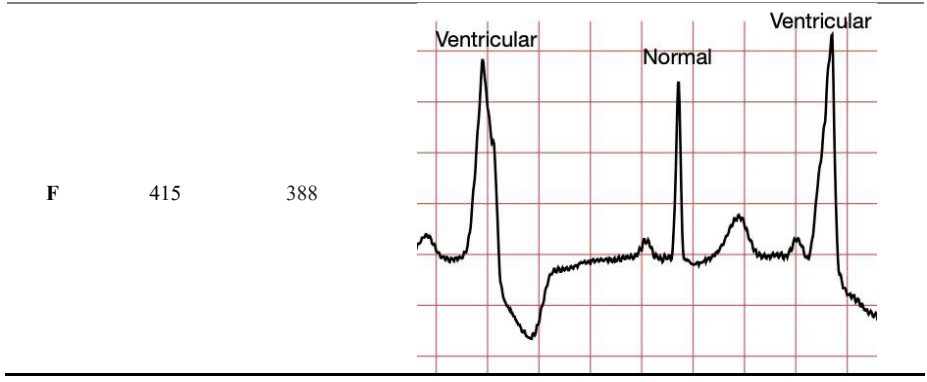 |

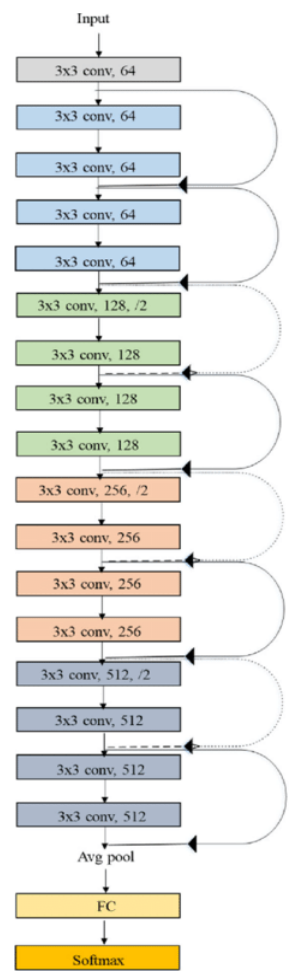

**Fig. A1**. Pretrained ResNet-18 architecture used in this study.